%% file: atw_arxiv.tex
\documentclass{article}

\usepackage{atw_preprint,times}
\input{math_commands.tex}

\usepackage{graphicx}
\usepackage{wrapfig}
\usepackage{placeins}
\usepackage{booktabs}
\usepackage{array}
\usepackage{tabularx}
\usepackage{colortbl}
\usepackage{soul}
\usepackage{hyperref}
\usepackage{url}
\input{visual_style.tex}

\newcolumntype{L}[1]{>{\raggedright\arraybackslash}p{#1}}
\newcolumntype{C}[1]{>{\centering\arraybackslash}p{#1}}
\newcolumntype{Y}{>{\centering\arraybackslash}X}
\newcommand{\method}{\textsc{ATW}}
\makeatletter
\newcommand{\minipagefigurecaption}{\def\@captype{figure}\caption}
\makeatother
\definecolor{ATWDraftYellow}{HTML}{FFF3B0}
\definecolor{ATWDraftBorder}{HTML}{D9A800}

\title{\raggedright\hyphenpenalty=10000\exhyphenpenalty=10000 Asking the World: Generalist Physical Reasoning through Agentic World Modeling~and~Probing}

\author{\normalfont
Shenxiang Zeng\textsuperscript{1}\thanks{Equal contribution.}\kern0.4em, Chen Yang\textsuperscript{1}\footnotemark[1]\kern0.4em, Peiyao Chen\textsuperscript{1},
Guohui Zhang\textsuperscript{1}, Jiansheng Fan\textsuperscript{1}\thanks{Corresponding authors: \href{mailto:fanjsh@tsinghua.edu.cn}{fanjsh@tsinghua.edu.cn}; \href{mailto:chwang@tsinghua.edu.cn}{chwang@tsinghua.edu.cn}.}\kern0.4em, Chen Wang\textsuperscript{1}\footnotemark[2]\\[3pt]
\multicolumn{1}{c}{\normalfont \textsuperscript{1}Tsinghua University}}

\iclrfinalcopy
\hypersetup{
 pdftitle={Asking the World: Generalist Physical Reasoning through Agentic World Modeling and Probing},
 pdfauthor={Shenxiang Zeng, Chen Yang, Peiyao Chen, Guohui Zhang, Jiansheng Fan, Chen Wang}
}
\begin{document}

\maketitle

\begingroup
\setlength{\intextsep}{-3pt}
\setlength{\abovecaptionskip}{4pt}
\setlength{\belowcaptionskip}{5pt}
\begin{figure}[!ht]
\centering
\includegraphics[width=0.90\linewidth]{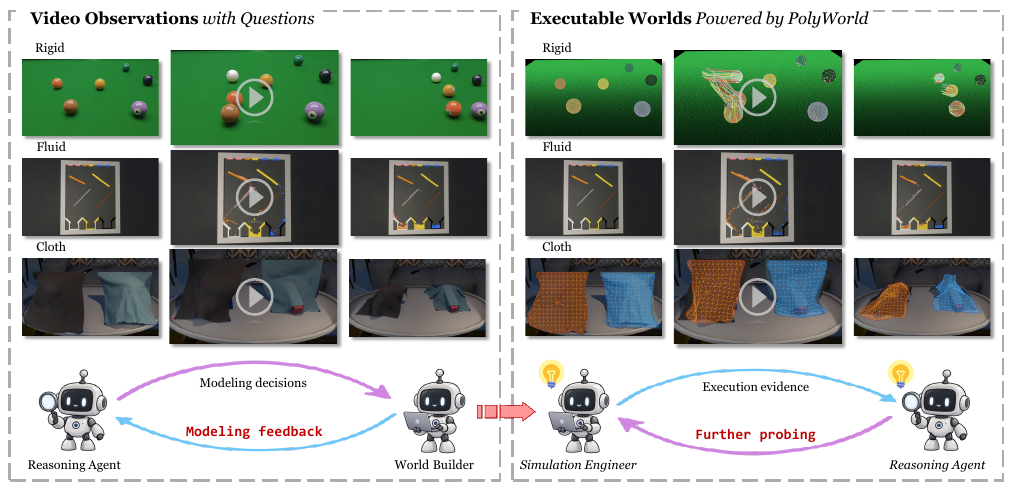}
\caption{\textbf{Physical reasoning through world modeling and probing.} Given videos and questions, \method{} builds and revises executable worlds, probes them with PolyWorld Engine, and reasons from execution evidence. Examples span rigid bodies, fluids, and cloth.}
\label{fig:teaser}
\end{figure}
\endgroup

\begin{abstract}
Physical reasoning from video requires inferring latent physical properties and dynamics beyond direct observation. Direct VLM inference remains unreliable on complex physical tasks without explicit modeling and validation, while predefined tool pipelines rely on task- and domain-specific priors that limit generalization across materials, dynamics, and reasoning tasks. We introduce \emph{Asking the World} (\method), a generalist agent that constructs and interrogates task-relevant executable worlds through two adaptive stages: \emph{World Modeling} calibrates a world from video, while \emph{World Probing} queries, simulates, and intervenes on it to obtain question-relevant evidence. Rather than prescribing the operations in either stage, \method{} determines how to model and probe according to the scene and question. We develop \emph{PolyWorld Engine}, a lightweight and highly programmable Warp-based multiphysics simulator for constructing and probing worlds with rigid bodies, soft bodies, cloth, ropes, fluids, and their coupled interactions. CEM-based system identification recovers task-relevant dynamics during World Modeling. The resulting world becomes an active workspace for question-directed physical experiments rather than a predetermined downstream tool. We evaluate \method{} on CLEVRER, ContPhy, and three real-world scenarios. Using Gemini-3-Flash as its base VLM, \method{} achieves 80.82\% overall per-question accuracy on CLEVRER, improving direct Gemini-3-Flash by 46.50 points, GPT-5.5 by 13.58 points, and PhysMind by 8.27 points. On ContPhy, it reaches 70.56\% overall accuracy, surpassing Gemini-3-Flash by 28.10 points and GPT-5.5 by 3.53 points. Across the three real-world scenarios, \method{} achieves 71.67\% accuracy, 28.33 points above GPT-5.5. These results establish agentic world modeling and probing as an effective, execution-grounded approach to generalist physical reasoning. Project page: \url{https://asking-the-world.github.io/}
\end{abstract}

\clearpage
\input{sections/introduction}
\input{sections/related_work}
\input{sections/method_plan}

\input{sections/experiments_plan_before_tables}

\begin{table}[t]
\caption{CLEVRER accuracy (\%) on a validation subset of 1,000 videos. Explanatory questions identify causes, while predictive and counterfactual questions concern future and altered outcomes. Each category reports per-question and per-option accuracy; a question is correct only when all associated options are correct. Bold and underlined values indicate the best and second-best results in each column, respectively.}
\label{tab:clevrer-main}
\centering
\footnotesize
\setlength{\tabcolsep}{2.7pt}
\renewcommand{\arraystretch}{1.08}
\begin{tabular*}{\textwidth}{@{\hspace{6pt}\extracolsep{\fill}}l*{8}{c}@{\hspace{6pt}}}
\toprule
Method
& \multicolumn{2}{c}{Explanatory}
& \multicolumn{2}{c}{Predictive}
& \multicolumn{2}{c}{Counterfactual}
& \multicolumn{2}{c}{Overall} \\
\cmidrule(lr){2-3}\cmidrule(lr){4-5}\cmidrule(lr){6-7}\cmidrule(lr){8-9}
& per ques. & per opt.
& per ques. & per opt.
& per ques. & per opt.
& per ques. & per opt. \\
\midrule
\rowcolor{ATWTableBand}[6pt][6pt]
\multicolumn{9}{@{\hspace{6pt}}l@{\hspace{6pt}}}{\textit{Baselines}} \\
Random & 7.19 & 49.33 & 25.40 & 49.71 & 9.75 & 50.02 & 11.26 & 49.69 \\
Blind Gemini-3-Flash & 27.00 & 64.64 & 24.68 & 48.92 & 10.07 & 50.13 & 19.21 & 56.31 \\
\midrule
\rowcolor{ATWTableBand}[6pt][6pt]
\multicolumn{9}{@{\hspace{6pt}}l@{\hspace{6pt}}}{\textit{Foundation VLMs}} \\
Qwen3-VL-235B-A22B & 53.19 & 75.84 & 40.40 & 62.12 & 22.17 & 61.79 & 37.52 & 67.92 \\
GLM-4.6V & 39.33 & 73.75 & 63.64 & 66.74 & 24.31 & 58.70 & 36.68 & 66.02 \\
Gemini-3-Flash & 50.32 & 74.66 & 42.71 & 53.46 & 16.63 & 58.38 & 34.32 & 64.97 \\
Gemini-3.1-Pro & 60.90 & 79.17 & 61.76 & 71.28 & 25.37 & 63.50 & 45.47 & 71.06 \\
GPT-4o & 36.76 & 70.48 & 35.64 & 54.98 & 15.57 & 56.54 & 27.29 & 62.44 \\
GPT-5.5 & \underline{77.21} & \underline{89.55} & \underline{85.71} & \underline{90.69} & 51.33 & 76.74 & 67.24 & 83.66 \\
\midrule
\rowcolor{ATWTableBand}[6pt][6pt]
\multicolumn{9}{@{\hspace{6pt}}l@{\hspace{6pt}}}{\textit{Training-Based Methods}} \\
VideoRFT & 16.66 & 61.98 & 39.25 & 47.40 & 15.35 & 54.98 & 19.74 & 57.28 \\
Video-R1 & 32.79 & 71.14 & 41.56 & 46.25 & 19.62 & 59.10 & 28.43 & 63.08 \\
VideoThinker-R1 & 17.53 & 63.60 & 41.56 & 42.50 & 15.14 & 53.71 & 20.37 & 56.92 \\
Chain-of-Frames & 44.71 & 77.26 & 84.42 & 87.88 & 33.48 & 70.84 & 46.21 & 75.29 \\
\midrule
\rowcolor{ATWTableBand}[6pt][6pt]
\multicolumn{9}{@{\hspace{6pt}}l@{\hspace{6pt}}}{\textit{Training-Free Methods}} \\
VideoAgent & 34.42 & 61.55 & 14.29 & 48.56 & 7.57 & 51.61 & 19.39 & 55.63 \\
STAR & 55.58 & 78.62 & 24.24 & 57.14 & 7.57 & 53.47 & 29.46 & 64.75 \\
PhysMind & 76.97 & 87.38 & 66.96 & 82.32 & \underline{70.58} & \underline{88.08} & \underline{72.55} & \underline{87.22} \\
\textbf{\method{} (Ours)} & \textbf{83.87} & \textbf{92.31} & \textbf{88.60} & \textbf{91.41} & \textbf{75.16} & \textbf{90.29} & \textbf{80.82} & \textbf{91.28} \\
\bottomrule
\end{tabular*}
\end{table}

\begin{table}[t]
\caption{ContPhy accuracy (\%) over 600 videos and 1,950 questions. Property questions test physical attributes, predictive and counterfactual questions test future and intervened outcomes, and goal-driven questions test action selection toward a target state. Bold and underlined values indicate the best and second-best results in each column, respectively.}
\label{tab:contphy-main}
\centering
\small
\setlength{\tabcolsep}{4pt}
\renewcommand{\arraystretch}{1.10}
\begin{tabular*}{\textwidth}{@{\hspace{6pt}\extracolsep{\fill}}l*{5}{c}@{\hspace{6pt}}}
\toprule
Method & Property & Predictive & Counterfactual & Goal-driven & Overall \\
\midrule
\rowcolor{ATWTableBand}[6pt][6pt]
\multicolumn{6}{@{\hspace{6pt}}l@{\hspace{6pt}}}{\textit{Baselines}} \\
Random & 40.53 & 31.50 & 12.92 & 13.90 & 28.36 \\
Blind Gemini-3-Flash & 47.47 & 41.26 & 23.16 & 29.34 & 37.90 \\
\midrule
\rowcolor{ATWTableBand}[6pt][6pt]
\multicolumn{6}{@{\hspace{6pt}}l@{\hspace{6pt}}}{\textit{Foundation VLMs}} \\
Qwen3-VL-235B-A22B & 58.80 & 46.54 & 30.51 & 27.41 & 45.03 \\
Gemini-3-Flash & 60.00 & 43.90 & 23.16 & 22.39 & 42.46 \\
GPT-5.5 & \underline{75.47} & \underline{50.81} & \textbf{69.49} & \textbf{69.11} & \underline{67.03} \\
\midrule
\rowcolor{ATWTableBand}[6pt][6pt]
\multicolumn{6}{@{\hspace{6pt}}l@{\hspace{6pt}}}{\textit{Training-Based Methods}} \\
Video-R1-7B & 43.47 & 34.15 & 11.14 & 15.44 & 29.95 \\
Chain-of-Frames-8B & 50.27 & 48.17 & 10.24 & 20.85 & 36.62 \\
\midrule
\rowcolor{ATWTableBand}[6pt][6pt]
\multicolumn{6}{@{\hspace{6pt}}l@{\hspace{6pt}}}{\textit{Training-Free Methods}} \\
\textbf{\method{} (Ours)} & \textbf{81.73} & \textbf{62.20} & \underline{63.03} & \underline{67.18} & \textbf{70.56} \\
\bottomrule
\end{tabular*}
\end{table}

\input{sections/experiments_plan_after_tables}
\input{sections/endmatter_plan}

\bibliography{iclr2027_conference}
\bibliographystyle{iclr2027_conference}

\clearpage
\appendix
\input{sections/appendix_implementation}
\input{sections/appendix_benchmarks}
\input{sections/appendix_results}
\input{sections/appendix_prompts}

\end{document}

%% file: math_commands.tex
\usepackage{amsmath,amsfonts,bm}

\def\eqref#1{equation~\ref{#1}}
\def\1{\bm{1}}

\DeclareMathAlphabet{\mathsfit}{\encodingdefault}{\sfdefault}{m}{sl}
\SetMathAlphabet{\mathsfit}{bold}{\encodingdefault}{\sfdefault}{bx}{n}

%% file: visual_style.tex
\colorlet{ATWCite}{cyan}
\colorlet{ATWLink}{magenta}
\definecolor{ATWTableBand}{HTML}{F0F0F0}
\hypersetup{
  colorlinks=true,
  citecolor=ATWCite,
  linkcolor=ATWLink,
  urlcolor=ATWLink,
  filecolor=ATWLink,
  pdfborder={0 0 0}
}

%% file: sections/introduction.tex
\section{Introduction}

Physical questions about latent properties, future outcomes, and counterfactual events often require evidence beyond the observed video. A capable system must therefore construct a task-relevant account of scene dynamics and obtain physical evidence from it. Doing so across different materials, dynamics, and question types is the central ambition of generalist physical reasoning, from rigid-body collisions in CLEVRER to continuum phenomena in ContPhy~\citep{yi2020clevrer,zheng2024contphy}.

Existing approaches largely follow two paradigms. VLM-based methods exploit broad visual and semantic priors, either through direct inference or task-specific adaptation~\citep{chow2025physbench,wang2025videorft,feng2025videor1}. Yet recent evaluations continue to expose failures in physical consistency and causal evolution across multimodal and generative video models~\citep{puyin2026quantiphy,bansal2024videophy,meng2024worldsim,li2025worldmodelbench}. Physics-grounded methods instead recover dynamics, call specialized tools, or execute simulators~\citep{ding2021dynamic,li2023pacnerf,xie2024physgaussian,zhang2024physdreamer,fan2025star,cherian2026llmphy,yang2026physmind}. Although they provide explicit physical evidence, they usually commit in advance to a particular representation, physical model, or inference procedure, limiting adaptation across materials and questions. What remains missing is a general agent that can construct an appropriate executable world for the scene and determine how to interrogate it for the question at hand.

Recent agents show that tools, generated programs, and execution can support iterative reasoning rather than only terminal prediction~\citep{yao2023react,gupta2023visprog,suris2023vipergpt,liang2023codepolicies,yin2026viga}. Physical question answering, however, has no fully observed target scene against which every update can be compared; the agent must decide for itself what hidden mechanisms and interventions matter. We therefore introduce \emph{Asking the World} (\method), a generalist physical reasoning agent that organizes its reasoning into two stages. \emph{World Modeling} constructs a task-relevant executable world by identifying relevant entities, selecting physical representations, and recovering latent dynamics. \emph{World Probing} queries, simulates, and intervenes on that world to gather evidence for the associated questions. Rather than prescribing the operations in either stage, \method{} determines how to model and probe according to the scene and question. In a nutshell, \method{} asks the world by building and probing it (Figure~\ref{fig:teaser}). This process is enabled by \emph{PolyWorld Engine}, our lightweight and highly programmable multiphysics simulator built on Warp~\citep{warp2022}. It provides a unified environment for rigid bodies, soft bodies, cloth, ropes, fluids, and their coupled interactions, which \method{} can instantiate and modify on demand. CEM-based system identification supports World Modeling by recovering task-relevant parameters from visual evidence. Together, these capabilities turn simulation from a predetermined downstream tool into an active workspace for constructing and interrogating physical explanations.

Without task-specific training, \method{} applies the same agentic formulation across rigid-body, continuum, and real-world settings. It achieves 80.82\% overall per-question accuracy on CLEVRER, improving Gemini-3-Flash by 46.50 points, GPT-5.5 by 13.58 points, and PhysMind by 8.27 points. On ContPhy, it reaches 70.56\%, surpassing Gemini-3-Flash by 28.10 points and GPT-5.5 by 3.53 points; across three real-world scenarios, it achieves 71.67\% accuracy, 28.33 points above GPT-5.5. These results establish agentic world modeling and probing, supported by a general multiphysics engine, as an effective path toward generalist physical reasoning.

In summary, our contributions are threefold:
\begin{itemize}
  \setlength{\itemsep}{2pt}
  \setlength{\parskip}{0pt}
  \setlength{\parsep}{0pt}
  \item \textbf{Agentic physical reasoning.} We introduce \method{}, a generalist physical reasoning agent that organizes reasoning into World Modeling, which constructs task-relevant executable worlds, and World Probing, which conducts question-directed queries, simulations, and interventions.
  \item \textbf{Generalist multiphysics execution.} We develop PolyWorld Engine, a lightweight and highly programmable environment that unifies rigid bodies, soft bodies, cloth, ropes, fluids, and their coupled interactions for on-demand world construction, intervention, and validation.
  \item \textbf{Broad empirical validation.} We demonstrate substantial improvements over direct VLM reasoning and strong baselines across CLEVRER, ContPhy, and three real-world scenarios, spanning multiple materials, dynamical systems, and reasoning tasks.
\end{itemize}

%% file: sections/related_work.tex
\section{Related Work}

\paragraph{Visual Physical Reasoning.}
Object-centric models learn structured representations and dynamics~\citep{chen2021dcl,ding2021dynamic,wu2023slotformer,li2025slotpi}. VLM methods use targeted training, memories, perception modules, spatiotemporal tools, or simulator search~\citep{balazadeh2025pcbs,wang2025videorft,feng2025videor1,ghazanfari2025chainofframes,chow2025physbench,fan2024videoagent,fan2025star,cherian2026llmphy}, while complementary benchmarks assess physical consistency and world-model behavior in generated videos~\citep{bansal2024videophy,meng2024worldsim,li2025worldmodelbench}. Their physics nevertheless remains implicit or procedurally fixed; \method{} instead lets execution feedback select what to model and test.

\paragraph{Executable Physical Worlds.}
Differentiable simulators and video-conditioned models optimize parameters within specified dynamics~\citep{howell2022dojo,ding2021dynamic,li2023pacnerf}, while black-box approaches search simulator parameters from trajectory error~\citep{cherian2026llmphy}. Video-to-simulation methods reconstruct geometry and physical properties~\citep{xie2024physgaussian,zhang2024physdreamer,chen2025vid2sim,zhao2025physsplat}, and executable-world methods use reconstructed dynamics for question answering~\citep{yang2026physmind}. Methods model dynamics from images and text by selecting scene abstractions and simulators~\citep{omahony2025vdaworld}. These systems typically fix representations or material families, or infer dynamics from static inputs; \method{} instead builds task-sufficient, video-calibrated worlds with heterogeneous materials and coupled interactions.

\paragraph{Agents Reasoning through Engines.}
Language and vision agents interleave reasoning with actions, tools, or generated programs~\citep{yao2023react,gupta2023visprog,suris2023vipergpt,liang2023codepolicies}. Agentic reconstruction combines visual models, code generation, execution, and feedback to construct explicit scenes~\citep{yao2025cast,yin2026viga}; related agents use 3D programs for spatial reasoning~\citep{luo2026pyspatial,chen2026gca} or interactive environments for manipulation planning~\citep{liu2026simpact,xu2025etot}. These agents target observable scenes or prespecified goals, whereas physical questions may concern hidden, future, or counterfactual dynamics. \method{} therefore makes world construction question-conditioned and chooses which physical hypotheses to instantiate and probe.

%% file: sections/method_plan.tex
\begin{figure}[t]
\centering
\includegraphics[width=\linewidth]{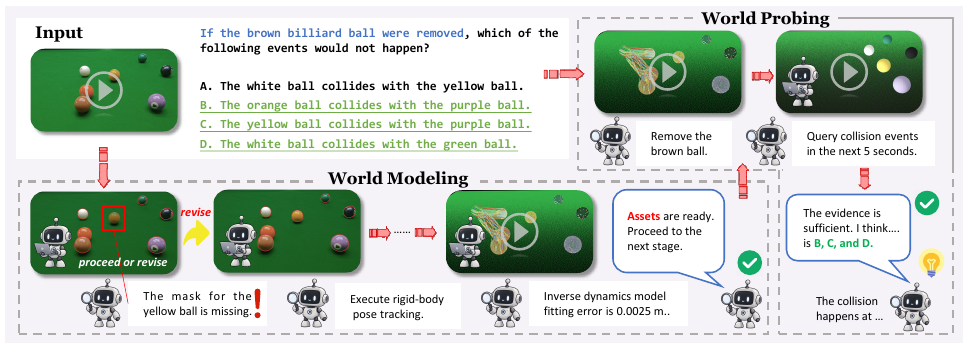}
\caption{\textbf{ATW system overview.} A billiards counterfactual illustrates the one-way transition from World Modeling to World Probing. The modeling agent corrects a missing object mask, tracks object poses, and fits dynamics before committing the executable world. The probing agent then removes the brown ball, queries collision events, and answers from the resulting evidence.}
\label{fig:overview}
\end{figure}

\section{Method}
\label{sec:method}

\method{} turns physical reasoning into interaction with an executable world. Given a video and its associated questions, the agent first constructs a task-sufficient physical world and then uses that world to conduct question-directed experiments. These are two consecutive stages: \emph{World Modeling} revises perceptual and physical assumptions until it commits a world, after which \emph{World Probing} adapts experiments within that world. Figure~\ref{fig:overview} illustrates this progression. We first describe the two-stage agentic framework, then present the PolyWorld Engine that makes executable worlds programmable by the agent, and finally explain how video evidence grounds and calibrates those worlds.

\subsection{ATW: Two-Stage Agentic Physical Reasoning}

Let $\mathcal{V}$ denote an input video and $\mathcal{Q}=\{q_i\}_{i=1}^{N}$ its physical questions. Rather than reconstructing every visible detail, \method{} seeks an executable world $\mathcal{W}^{\star}$ containing the entities, physical representations, states, and latent parameters needed by these questions. We denote the stage-specific agent policies by $\mathcal{A}_{\mathrm{M}}$ and $\mathcal{A}_{\mathrm{P}}$. The former constructs the world, while the latter obtains evidence $\mathcal{E}_i$ for each question and produces the answer:
\begin{equation}
\mathcal{W}^{\star}=\mathcal{A}_{\mathrm{M}}(\mathcal{V},\mathcal{Q}),
\qquad
\mathcal{E}_i=\mathcal{A}_{\mathrm{P}}(\mathcal{W}^{\star},q_i),
\qquad
\hat a_i=\operatorname{Answer}(q_i,\mathcal{E}_i).
\label{eq:two-stage-atw}
\end{equation}
Both $\mathcal{A}_{\mathrm{M}}$ and $\mathcal{A}_{\mathrm{P}}$ are adaptive agents rather than fixed workflows: within its stage, each agent chooses its next operation from the current state, prior tool results, and execution feedback.

\paragraph{World Modeling.}
The modeling agent decides what must be represented before deciding how to recover it. It selects question-relevant entities, invokes visual tools to establish their geometry and motion, assigns a physical representation to each entity, and identifies parameters whose values must be inferred. The agent can inspect intermediate masks, tracks, reconstructions, and simulated trajectories; a mismatch can trigger another observation, a corrected object binding, a different physical representation, or renewed parameter fitting. This feedback remains internal to World Modeling. The stage terminates by committing $\mathcal{W}^{\star}$, including object identities, geometry, initial states, physical models, and fitted parameters.

\paragraph{World Probing.}
The probing agent treats $\mathcal{W}^{\star}$ as a fixed base world and decides which physical experiment will resolve each question. It can read an inferred property, continue the unmodified dynamics, instantiate an intervened branch, inspect events or trajectories, and compare outcomes across branches. Thus property questions can query the fitted world directly, predictive questions can extend its trajectory, and counterfactual or goal-driven questions can test edited worlds. Execution feedback may change the next probe or correct its intervention target, but it does not reopen World Modeling or alter the committed base world. Probing ends when the accumulated evidence supports an answer or the interaction budget is exhausted.

\begin{figure}[t]
\centering
\includegraphics[width=\linewidth]{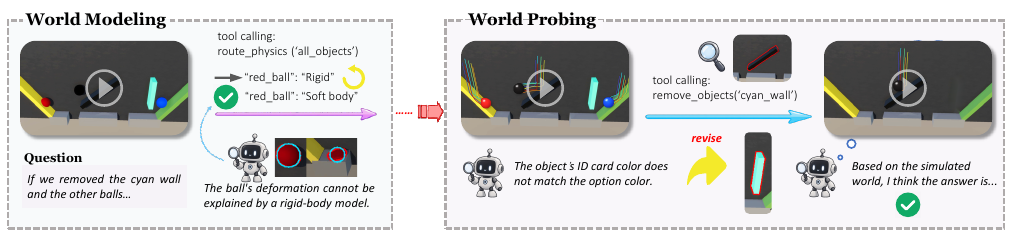}
\caption{\textbf{Stage-local feedback and correction.} Within World Modeling, observed deformation causes the agent to replace a rigid-body hypothesis with a soft-body model. Within World Probing, execution feedback reveals an incorrect intervention target and causes the agent to probe the intended cyan wall. Feedback therefore changes the agent's decisions within each stage.}
\label{fig:adaptive-reasoning}
\end{figure}

\subsection{PolyWorld Engine: Agent-Operable Multiphysics Worlds}

PolyWorld Engine is our lightweight and highly programmable multiphysics simulator built on Warp~\citep{warp2022}. It places rigid bodies, soft bodies, cloth, ropes, and fluids in a common executable scene and supports their coupled interactions. Each physical system retains the state variables and dynamics appropriate to its material, while the shared scene allows the agent to compose heterogeneous systems when required by the observed interaction. Reusable scene data, simulation buffers, and explicit parameterization make repeated fitting and branched rollouts practical within an agent trajectory.

PolyWorld exposes physics through semantic operations rather than requiring the agent to manipulate low-level simulator code. Table~\ref{tab:polyworld-interface} summarizes the interface used across the two stages. The available operations are conditioned on the instantiated physical systems, and every result records its source world, objects, temporal scope, and execution status. This provenance lets the agent revisit earlier evidence without rerunning an experiment or confusing observations from different branches.

\begin{table}[t]
\caption{\textbf{PolyWorld as an agent-facing semantic interface.} High-level operations connect agent decisions to physical execution while preserving explicit world and evidence provenance. Concrete tool signatures and backend-specific details are provided in Appendix~\ref{app:implementation}.}
\label{tab:polyworld-interface}
\centering
\small
\setlength{\tabcolsep}{4pt}
\renewcommand{\arraystretch}{1.08}
\begin{tabularx}{\linewidth}{@{}L{0.19\linewidth}L{0.39\linewidth}>{\raggedright\arraybackslash}X@{}}
\toprule
Interface role & Representative capabilities & Returned state or evidence \\
\midrule
World construction & Bind objects, assign physical models, initialize states, fit parameters & Identities, geometry, model assignments, fitted dynamics \\
World inspection & Inspect frames or rollouts, read properties, query states & Visual observations, properties, trajectories, events \\
World experimentation & Continue dynamics, intervene, branch rollouts, compare worlds & Predictive and counterfactual outcomes with temporal scope \\
Execution control & Describe operations, revisit evidence, recover from failure, terminate & Tool constraints, execution status, retained evidence \\
\bottomrule
\end{tabularx}
\end{table}

The committed world is reusable but not destructively edited during probing. Predictive rollouts continue the base world, whereas interventions create branches that inherit its fitted parameters and initial conditions. Multiple questions and candidate interventions can therefore share the same calibrated world, while their trajectories and events remain separately addressable. This design turns simulation into a persistent workspace for physical inquiry rather than a terminal tool invocation.

\subsection{Grounding and Calibrating Executable Worlds}

World Modeling converts video evidence into a world that is physically executable and sufficient for the questions at hand. The agent obtains object masks, motion tracks, and scene geometry from visual tools, and retains explicit bindings between referenced entities and objects in the executable scene. These observations play complementary roles: masks delimit objects and their visible deformation, tracks constrain temporal motion, and geometry establishes spatial configuration and initial conditions. The agent requests or refines only the evidence required by the selected representation, avoiding a mandatory reconstruction pipeline shared by all scenes.

For each instantiated physical representation $h$, PolyWorld maps an initial state $s_0$ and parameters $\boldsymbol{\theta}$ to a simulated trajectory, from which a model-specific readout is compared with visual evidence. Rigid systems use tracked positions and orientations; deformable systems additionally use shape, surface, centerline, or occupancy observations as appropriate. We fit the exposed parameters with the cross-entropy method (CEM). At iteration $k$, CEM evaluates bounded parameter samples under the corresponding trajectory discrepancy and uses the lowest-loss elite set to update its diagonal Gaussian proposal:
\begin{equation}
\begin{aligned}
\boldsymbol{\mu}_{k+1}
&= \alpha\boldsymbol{\mu}_k+(1-\alpha)\widehat{\boldsymbol{\mu}}_k,\\
\boldsymbol{\sigma}_{k+1}
&= \max\!\left(\boldsymbol{\sigma}_{\min},
\alpha\boldsymbol{\sigma}_k+(1-\alpha)\widehat{\boldsymbol{\sigma}}_k\right),
\end{aligned}
\label{eq:cem-update}
\end{equation}
where $\widehat{\boldsymbol{\mu}}_k$ and $\widehat{\boldsymbol{\sigma}}_k$ are the elite statistics, $\alpha$ controls smoothing, and $\boldsymbol{\sigma}_{\min}$ preserves exploration. Fitting progressively incorporates longer observation windows so that later interactions constrain the recovered dynamics (system-identification details: Appendix~\ref{app:warp-cem}).

Calibration also tests whether the current world hypothesis is adequate. Persistent motion or shape discrepancies can indicate that the error lies in the representation or object binding rather than its numerical parameters. The modeling agent can therefore revise those choices and repeat fitting before committing $\mathcal{W}^{\star}$ as the shared physical basis for subsequent experiments.

%% file: sections/experiments_plan_before_tables.tex
\section{Experiments}

We evaluate \method{} on CLEVRER~\citep{yi2020clevrer}, ContPhy~\citep{zheng2024contphy}, and real-world videos against foundation VLMs and physical-reasoning methods. We also present qualitative results and ablations of adaptive reasoning, physical simulation, and system identification.

\subsection{Experimental Protocol}

\paragraph{Benchmarks and metrics.} CLEVRER~\citep{yi2020clevrer} tests explanatory, predictive, and counterfactual rigid-body reasoning. Its validation subset contains 1,000 videos, 4,280 questions, and 14,228 options; we report per-option and per-question accuracy, with the latter requiring every option to be correct. ContPhy~\citep{zheng2024contphy} covers physical-property, predictive, counterfactual, and goal-driven reasoning. On its 600 videos and 1,950 questions, we report category accuracy and a question-count-weighted overall score. We additionally report category-wise accuracy on three real-world scenarios totaling 60 videos and questions (evaluation details: Appendix~\ref{app:benchmarks}).

\paragraph{Baselines and controlled variants.} Baselines comprise foundation VLMs---Qwen3-VL-235B-A22B~\citep{bai2025qwen3vl}, GLM-4.6V~\citep{zai2025glm46v}, Gemini-3-Flash~\citep{google2025gemini3flash}, Gemini-3.1-Pro~\citep{google2026gemini31pro}, GPT-4o~\citep{openai2024gpt4o}, and GPT-5.5~\citep{openai2026gpt55}; training-based video reasoners---VideoRFT~\citep{wang2025videorft}, Video-R1~\citep{feng2025videor1}, VideoThinker-R1~\citep{wu2026videothinker}, and Chain-of-Frames~\citep{ghazanfari2025chainofframes}; and training-free methods---VideoAgent~\citep{fan2024videoagent}, STAR~\citep{fan2025star}, and PhysMind~\citep{yang2026physmind}. Controlled variants isolate structured perception, physical execution, and adaptive reasoning.

\paragraph{Implementation details.} \method{} uses Gemini-3-Flash and PolyWorld Engine, without training or fine-tuning on either benchmark. Gemini baselines receive the video; GPT, Qwen, and GLM receive eight uniformly sampled frames (implementation details: Appendix~\ref{app:models}; prompts: Appendix~\ref{app:system-prompts}).

\subsection{Main Results}

%% file: sections/experiments_plan_after_tables.tex
\paragraph{CLEVRER.} Table~\ref{tab:clevrer-main} shows that \method{} achieves 80.82\% overall per-question accuracy, outperforming Gemini-3-Flash by 46.50 percentage points, GPT-5.5 by 13.58 points, and PhysMind, the strongest training-free baseline, by 8.27 points. \method{} leads all three categories with 83.87\% explanatory, 88.60\% predictive, and 75.16\% counterfactual accuracy. Its largest gain over GPT-5.5 is on counterfactual questions (+23.83 points), where it also exceeds PhysMind by 4.58 points.

\paragraph{ContPhy.} Table~\ref{tab:contphy-main} evaluates continuous-media and diverse-material tasks. \method{} achieves 70.56\% overall accuracy, 28.10 percentage points above Gemini-3-Flash and 3.53 points above GPT-5.5; it also outperforms Gemini-3-Flash in each of the four categories. Its gains over GPT-5.5 concentrate in physical-property and predictive reasoning, reaching 81.73\% and 62.20\% (+6.26 and +11.39 points). It nearly matches GPT-5.5 on goal-driven questions (a 1.93-point gap) but trails it by 6.46 points on counterfactual questions.

\par\medskip
\noindent
\begin{minipage}[c]{0.54\textwidth}
\textbf{Real-world evaluation.}
Across the three real-world tasks, \method{} achieves 85.00\%, 70.00\%, and 60.00\% per-question accuracy on billiards counterfactual, bouncing-ball prediction, and toy-car prediction, respectively, outperforming both GPT-5.5 and Gemini-3-Flash in every category. Its overall accuracy is 71.67\%, compared with 43.33\% for GPT-5.5 and 33.33\% for Gemini-3-Flash. Across these 60 videos, the gains show that executable-world reasoning extends to the three evaluated real-world settings despite appearance variation. The lower predictive-question score also reflects the greater difficulty of forecasting future trajectories in real-world scenes (detailed results: Appendix~\ref{app:real-world-results}).
\end{minipage}\hfill
\begin{minipage}[c]{0.42\textwidth}
\centering
\includegraphics[width=0.82\linewidth]{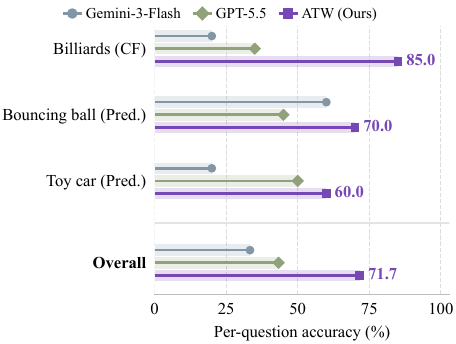}
\minipagefigurecaption{Real-world per-question accuracy (CF: counterfactual; Pred.: prediction).}
\label{fig:real-world-accuracy}
\end{minipage}
\par\medskip

\begin{figure}[t]
\centering
\includegraphics[width=\textwidth]{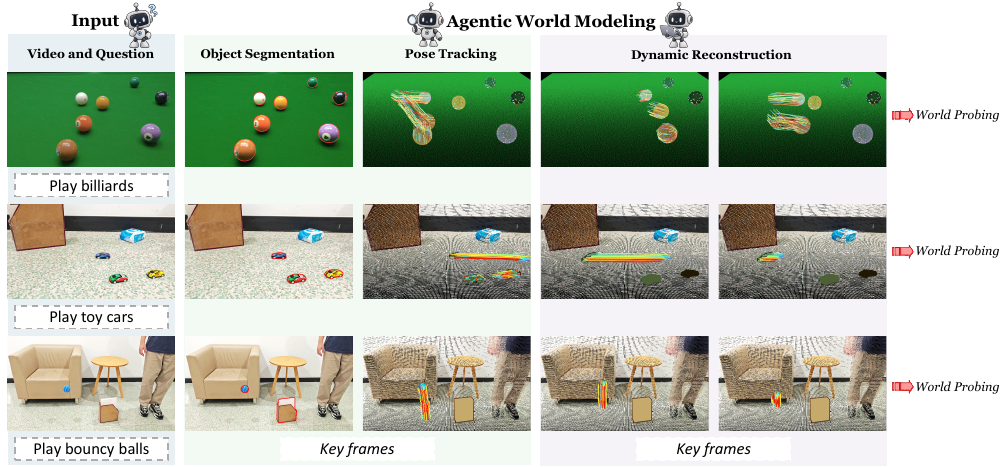}
\caption{Agentic world modeling from real-world videos of billiards, toy cars, and bouncing balls. Selected key frames show object segmentation, pose tracking, and dynamic reconstruction, producing executable worlds for subsequent probing.}
\label{fig:real-world-modeling}
\end{figure}

Figure~\ref{fig:real-world-modeling} visualizes the selected modeling steps across three real-world tasks. Segmentation identifies task-relevant objects, pose tracking preserves identities and motion across key frames, and dynamic reconstruction forms executable states. Together, they provide the geometry, trajectories, and interactions needed for targeted prediction and counterfactual probes.

\begin{figure}[t]
\centering
\includegraphics[width=\textwidth]{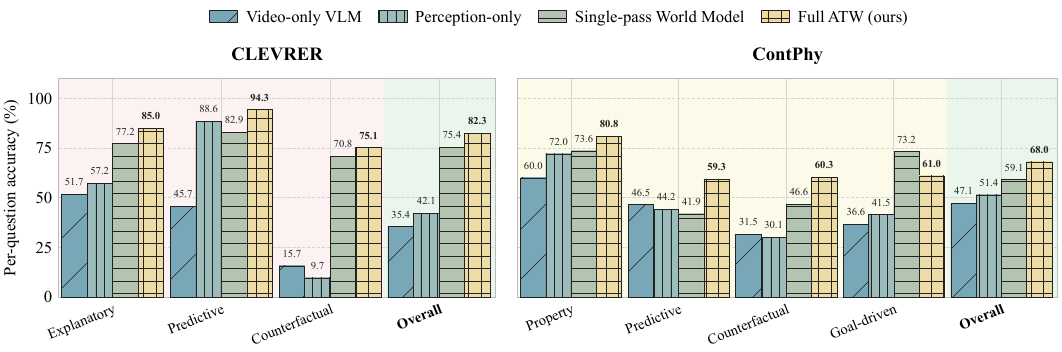}
\caption{Ablation of agentic world reasoning on the CLEVRER and ContPhy subsets. All four variants use Gemini-3-Flash. Bars and labels show per-question accuracy; overall scores aggregate all questions in each subset.}
\label{fig:agentic-reasoning-ablation}
\end{figure}

\FloatBarrier

\subsection{Ablation Studies}

We conduct ablation studies on CLEVRER and ContPhy subsets, each containing 100 scenes. CLEVRER includes 435 questions and 1,439 answer options, while ContPhy contains 325 questions. These studies examine the roles of executable worlds and adaptive reasoning, the effectiveness of physical simulation and system identification, and the applicability of \method{} across foundation models. Within each comparison, the data and evaluation protocol remain fixed so that performance changes isolate the component under study.

\paragraph{Agentic world reasoning.} Figure~\ref{fig:agentic-reasoning-ablation} compares four variants: Video-only VLM receives only the video and question; Perception-only (w/o Executable World) additionally receives \method{}\textquotesingle s structured detections, masks, tracks, and geometry, but neither builds nor executes a physical world; Single-pass World Model builds and executes a world once without feedback-driven revision or additional probing; and Full \method{} (ours) adaptively updates its world representation, physical parameters, and subsequent probes based on the question and accumulated feedback. Overall accuracy increases across these variants from 35.40\% to 42.07\%, 75.40\%, and 82.30\% on CLEVRER, and from 47.08\% to 51.38\%, 59.08\%, and 68.00\% on ContPhy. Structured perception improves video-only reasoning, while physical execution brings a larger gain. Relative to Perception-only, Single-pass World Model raises CLEVRER counterfactual accuracy by 61.08 percentage points, from 9.73\% to 70.81\%. Feedback-guided revision in Full \method{} then adds 6.90 and 8.92 points over Single-pass on CLEVRER and ContPhy, respectively. Despite some variation across individual reasoning categories, the overall trend across both datasets shows that executable worlds provide substantial gains, while feedback-guided revision brings further improvements (see Appendix~\ref{app:additional-ablations}).

\begin{figure}[htbp]
\centering
\begin{minipage}[t]{0.49\textwidth}
\centering
\includegraphics[width=\linewidth]{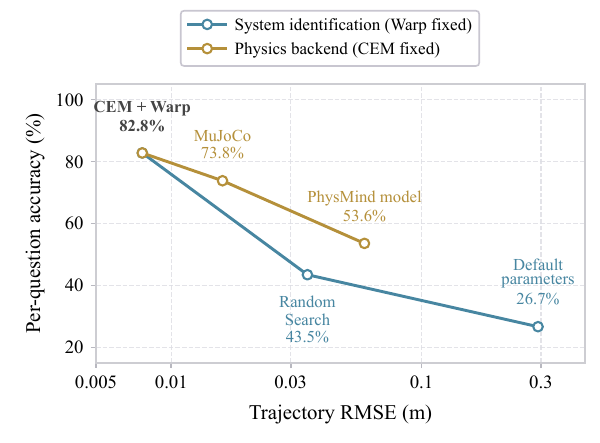}
\caption{System-identification and physics-backend ablations on CLEVRER. Overall per-question accuracy versus median trajectory RMSE (log scale). System identification fixes Warp; backend comparisons fix CEM. Both ablations share CEM + Warp. CEM and Random Search use equal simulation budgets.}
\label{fig:physical-grounding-ablation}
\end{minipage}\hfill
\begin{minipage}[t]{0.49\textwidth}
\centering
\includegraphics[width=\linewidth]{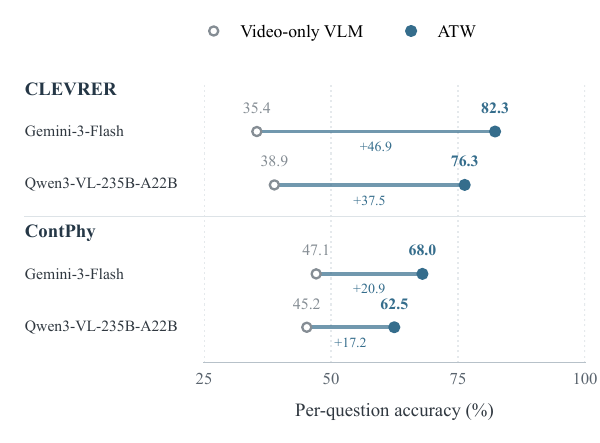}
\caption{Across foundation models. Overall per-question accuracy; open: Video-only VLM, filled: ATW. Values below the lines show gains in percentage points.}
\label{fig:backbone-generalization}
\end{minipage}
\end{figure}

\paragraph{System identification and physical backends.} Figure~\ref{fig:physical-grounding-ablation} evaluates both factors on the CLEVRER ablation subset. With Warp fixed and equal simulation budgets, CEM achieves a substantially lower median trajectory RMSE than Random Search, while raising per-question accuracy from 43.45\% to 82.76\% and per-option accuracy from 69.49\% to 93.19\%. With CEM fixed, Warp achieves the lowest trajectory error and the highest per-question accuracy among the PhysMind physical model~\citep{yang2026physmind}, MuJoCo~\citep{todorov2012mujoco}, and Warp. This paired design separates optimization quality from backend fidelity: stronger search improves the same simulator, while a more suitable backend improves reasoning under the same optimizer. Across both ablations, lower fitting error is consistently associated with stronger downstream reasoning, demonstrating that reliable physical grounding depends on both system identification and simulation (see Appendix~\ref{app:physical-grounding-details}).

\begin{samepage}
\paragraph{Generalization across foundation models.} Figure~\ref{fig:backbone-generalization} extends the Gemini-3-Flash evaluation to Qwen3-VL-235B-A22B, replacing the backbone in both world modeling and world probing. With Qwen, \method{} improves overall per-question accuracy over Video-only VLM from 38.85\% to 76.32\% on CLEVRER and from 45.23\% to 62.46\% on ContPhy. Across the two evaluated backbones, every paired comparison improves in the same direction, demonstrating backbone portability in both the tested rigid-body and continuous-physics settings.
\end{samepage}

\FloatBarrier

%% file: sections/endmatter_plan.tex
% Preserve the requested Conclusion and Appendix numbering.
\setcounter{section}{4}
\section{Conclusion}

We presented Asking the World (\method), a training-free framework for generalist physical reasoning through agentic world modeling and probing. Rather than treating simulation as a fixed downstream tool, \method{} constructs task-relevant executable worlds and uses observations and execution feedback to revise modeling decisions and guide question-directed probes. PolyWorld Engine supports diverse materials and coupled interactions, while system identification grounds world dynamics in visual evidence. Across CLEVRER, ContPhy, and three real-world scenarios, \method{} improves reasoning accuracy, while controlled ablations show complementary contributions from physical execution and feedback-guided adaptation. These findings suggest that actively building and interrogating executable worlds provides a promising path toward physical reasoning beyond direct visual inference.

\clearpage
\subsection*{AI use statement}

Generative AI tools, including large language models, were used to assist with implementing portions of the research code, drafting portions of the manuscript, and improving clarity and readability. They were not used to generate synthetic datasets or to formulate or prove mathematical claims. The authors reviewed and tested all AI-assisted code, verified the reported results, and take responsibility for the final content.

\subsection*{Ethics statement}

This work uses public benchmarks and author-recorded tabletop scenes and involves no human participants, personal data, or decisions about individuals. Because errors in perception or simulation can produce incorrect conclusions, the method should not be used in safety-critical settings without independent validation.

\subsection*{Reproducibility statement}

The paper describes the method, experimental protocol, metrics, baselines, and ablations. Appendices~\ref{app:implementation}--\ref{app:system-prompts} provide additional implementation context, benchmark and real-world protocols, supplementary experimental results, and system-prompt summaries to support interpretation and assessment of the reported findings.

%% file: sections/appendix_implementation.tex
\section{More Implementation Details}
\label{app:implementation}

\subsection{Models and Visual Foundation Modules}
\label{app:models}

\paragraph{Reasoning models.} Gemini-3-Flash~\citep{google2025gemini3flash} serves as the default backbone for both world modeling and world probing. The cross-backbone experiment replaces the model in both stages with Qwen3-VL-235B-A22B~\citep{bai2025qwen3vl}. These models select operations, interpret returned evidence, and produce answers. We do not train or fine-tune the reasoning models or visual modules on CLEVRER or ContPhy.

\paragraph{SAM 3.} SAM~3~\citep{carion2026sam3} supports concept-prompted segmentation of objects in images and videos. In ATW, the agent supplies object descriptions to obtain masks and associated video identities. These masks delimit the visual evidence used for object reconstruction and motion tracking, and connect image regions to retained scene entities. The agent can inspect the segmentation results and request another segmentation when the current masks do not adequately capture the objects relevant to modeling.

\paragraph{Track4World.} Track4World~\citep{lu2026track4world} provides dense 3D tracking in a shared world coordinate system. We use its Depth Anything~3 variant to obtain geometric observations and inter-frame correspondences from the input video. For dynamic rigid objects, correspondences initialize pose propagation, followed by point-cloud registration against the observed geometry. The resulting motion estimates are retained with object identities and confidence information, providing temporal evidence for fitting physical parameters and comparing simulated motion with observations.

\paragraph{Depth Anything 3.} Depth Anything~3~\citep{lin2025depthanything3} recovers scene geometry from visual inputs and serves as the geometric backbone of our Track4World configuration. Through this integration, the perception pipeline obtains camera intrinsics, camera poses, and dense 3D points in camera and world coordinates. These outputs provide the spatial reference for aligning reconstructed object geometry with the video. Geometry confidence and validity information support the selection of usable observations for downstream reconstruction and tracking.

\paragraph{SAM 3D Objects.} SAM~3D Objects~\citep{chen2026sam3d} reconstructs object geometry from an image and an object mask. ATW applies it to selected reference observations to obtain canonical object geometry, which is aligned with the recovered scene geometry. For rigid objects, this reference shape is retained while subsequent frames update the object's pose. Keeping geometry and motion separate provides a consistent object representation for tracking, physical modeling, and visualization of the reconstructed world.

\paragraph{FoundationPose.} FoundationPose~\citep{wen2024foundationpose} estimates and refines the 6D pose of an object using its geometry and visual observations. In our implementation, it refines the reference-frame alignment between the reconstructed object and the observed scene. This pose anchors the canonical object geometry in the recovered coordinate system. Subsequent dynamic poses use Track4World correspondences and point-cloud registration, while static objects retain their aligned reference pose. FoundationPose therefore supplies the initial geometric alignment used by temporal tracking.

\subsection{Warp-Based Simulation and CEM Fitting}
\label{app:warp-cem}

\paragraph{Shared simulation interface.} PolyWorld Engine uses Warp~\citep{warp2022} to execute physical models for rigid bodies, soft bodies, cloth, ropes, and fluids. Each model receives scene geometry, initial conditions, and a candidate parameter vector, and produces a simulated trajectory. Warp kernels accelerate the forward dynamics on the GPU, while reusable scene data and simulation buffers support repeated candidate evaluation. The fitting interface expresses this process as
\begin{equation}
 \widehat Y_{1:T}(\theta)=\mathcal O_h\!\left(\mathcal F_h(s_0,\theta;T)\right),
\label{eq:app-forward}
\end{equation}
where $h$ denotes the physical representation, $s_0$ its initial state, $\mathcal F_h$ the forward simulator, and $\mathcal O_h$ the readout that converts simulated states into quantities comparable to visual observations.

\paragraph{Observation-driven fitting.} System identification fits the parameters exposed by the instantiated physical model to the retained visual evidence. These parameters describe motion, material response, contact, or boundary conditions, depending on the system. Across the implementations, the fitting objective can be expressed as
\begin{equation}
 \theta^{\star}=\arg\min_{\theta\in\Theta_h}
 \mathcal L_h(\theta),\qquad
 \mathcal L_h(\theta)=D_h\!\left(\widehat Y_{1:T}(\theta),Y_{1:T}\right)
 +\lambda_h R_h(\theta),
\label{eq:app-fitting}
\end{equation}
where $Y_{1:T}$ is the observed evidence, $D_h$ measures the corresponding motion or shape discrepancy, and $\Theta_h$ specifies admissible parameters. $R_h$ represents a parameter prior when used, with $\lambda_h=0$ otherwise. Observation validity and confidence determine which measurements contribute to the discrepancy. This formulation allows each physical model to use the evidence that directly describes its dynamics.

\paragraph{Evidence across physical systems.} For \emph{rigid bodies}, fitting compares simulated and tracked positions and rotations to recover motion and contact-response parameters. For \emph{soft bodies}, it combines center motion with directional deformation, linking material response to observed changes in shape. For \emph{cloth}, it compares simulated surface points with tracked trajectories to fit motion, stiffness, damping, and boundary conditions. For \emph{ropes}, it uses centerline geometry and the trajectories of connected objects to fit constrained dynamics. For \emph{fluids}, it compares simulated spatial distributions with observed fluid regions to fit flow behavior and source conditions, including relative density and emission parameters. These observation choices adapt the shared fitting procedure to the representation of each system.

\paragraph{CEM search.} CEM maintains a diagonal Gaussian proposal over the fitted parameters. At iteration $k$, it samples a population, projects candidates onto their bounds, and selects the lowest-loss candidates as an elite set:
\begin{equation}
 \theta_k^{(j)}=\Pi_{\Theta_h}(\mu_k+\sigma_k\odot\epsilon_j),
 \quad\epsilon_j\sim\mathcal N(0,I),\qquad
 \mathcal E_k=\operatorname{TopK}_{\mathrm{lowest}\ \mathcal L_h}
 \{\theta_k^{(j)}\}_{j=1}^{N}.
\label{eq:app-cem}
\end{equation}
Here $N$ is the population size and $K$ is the elite count. The elite mean and standard deviation update the proposal through the smoothed rule in Equation~\ref{eq:cem-update}, with a variance floor maintaining exploration. The search reuses promising candidates and progressively extends the fitting horizon so that later interactions contribute to parameter fitting. Model-specific parameterizations and readouts connect this common search procedure to the different forward simulators.

\paragraph{Reusable fitted worlds.} The fitted parameters are retained with the modeled geometry and initial conditions, yielding an executable world that can be replayed and queried. World probing uses these assets to continue the dynamics or execute an intervention, such as removing an object or changing an exposed physical condition. The same forward model therefore connects parameter fitting with subsequent question-conditioned experiments.

\subsection{Query-Conditioned Physical Rollout Tools}
\label{app:rollout-tools}

The probing agent receives the retained world assets, object identities, available capabilities, and the current physical question. Tool availability and temporal scope depend on the question and the instantiated world. The following signatures are pseudocode abstractions of general operations; each backend exposes the subset supported by its physical representation.

\paragraph{Asset and property queries.} \texttt{InspectWorld(objects, fields)} retrieves identities, geometry, visibility, and available capabilities. \texttt{ReadProperty(objects, property)} reads supported physical quantities from the fitted world, allowing property questions to use existing parameters without an additional rollout. These queries preserve the distinction between observed attributes and fitted physical properties.

\paragraph{Rollouts and interventions.} \texttt{Rollout(world, intervention, horizon)} executes an unedited or modified world and returns a rollout identifier. Predictive questions use the unedited world and read the interval after the observation cutoff. Counterfactual questions specify an intervention before execution, while goal-directed questions can test candidate interventions in separate worlds. Object removal is one supported edit; other edits are exposed according to backend capabilities. In the rigid implementation, removal applies from the modeled beginning of the replay. Fitted parameters of the retained objects are reused for the edited rollout.

\paragraph{Evidence readout.} \texttt{ReadEvents(rollout, objects, interval)} and \texttt{ReadTrajectory(rollout, object, interval)} retrieve time-localized outcomes. \texttt{ReadTimeline(rollout, interval)} organizes collision, entry, and exit events chronologically, retaining event times and participating object identities to support reasoning about their order. \texttt{QueryWorld(rollout, predicate, interval)} requests semantic conditions, such as contact, motion, or region occupancy, with the appropriate temporal aggregation. Readouts refer to an explicit world and time interval. Several queries can accompany a rollout request, and retained rollouts can be queried repeatedly.

\paragraph{Cross-world comparison.} \texttt{CompareWorlds(base, edited, objects)} compares selected objects across an original world and an intervened world. It aligns their trajectories at shared frames and reports the maximum center displacement for each object, together with the two event sequences and simulation horizons. These outputs show how an intervention changes object motion and event occurrence, providing evidence for comparing the consequences of alternative physical conditions.

\paragraph{Observation-based checks.} \texttt{InspectObservedMotion(object, interval)} retrieves motion-related measurements from the object's masks in the input video. The returned observations include image-space centroid positions, visible mask areas, and image-boundary contact over the requested interval. Together with the original frames, these measurements let the agent inspect where and when an object moves and compare the observed changes with its tracked or simulated trajectory.

\paragraph{Inspection and control.} \texttt{DescribeTool(tool)} provides an operation's inputs, parameter constraints, and conditions of use, helping the agent prepare its next call. \texttt{InspectFrames(times, object)} retrieves visual observations, while \texttt{InspectRollout(rollout, interval)} presents the simulated interval for visual inspection of motion and deformation alongside numerical readouts. \texttt{ReadEvidence(id)} revisits an earlier tool result, and \texttt{UpdateBindings(correction, reason)} revises question-to-object associations using accumulated evidence. Finally, \texttt{Finish(evidence)} ends probing and passes the evidence to answer generation. Tool returns retain their observation or simulation source, object and rollout identities, temporal coverage, and execution status.

\subsection{SAM 3 Agent for General-Purpose Video Segmentation}
\label{app:sam3-agent}

\paragraph{Object discovery and seed selection.} The SAM~3 agent combines VLM-based visual reasoning with SAM~3~\citep{carion2026sam3} segmentation to produce object-level mask tracks. Given a video and indexed frame samples, the VLM identifies distinct entities in the interaction workspace and generates discriminative descriptions and segmentation prompts. It selects a reference frame for each entity, prioritizing complete outlines, visibility, and identity clarity. SAM~3 then generates candidate masks, which the VLM inspects alongside the original image to select a mask covering the target while excluding neighboring objects and background.

\paragraph{Feedback-guided refinement and propagation.} When no candidate is accepted, the agent checks the target's visibility and can select another reference frame or propose alternative descriptions of the same entity. Candidate masks and rejection feedback guide these adjustments while preserving the target identity. Accepted seeds initialize bidirectional video-mask propagation in a shared multi-object state. The module returns indexed mask sequences, object-to-track associations, and temporal quality summaries for downstream reconstruction and motion tracking. This interface supports reusable object segmentation across scenes without requiring an object mesh or a rigid-motion assumption.

%% file: sections/appendix_benchmarks.tex
\section{Evaluation Benchmark Details}
\label{app:benchmarks}

\subsection{CLEVRER}
\label{app:clevrer}

CLEVRER~\citep{yi2020clevrer} is a video reasoning benchmark designed to study causal understanding of physical events. It pairs synthetic videos of interacting rigid objects with questions linking object identities, motion, and collisions. Objects vary in shape, color, and material appearance, allowing questions to refer to specific entities and their interactions. We evaluate \method{} on a validation subset containing 1,000 videos, 4,280 questions, and 14,228 answer options.

\paragraph{Explanatory questions.} These questions ask which earlier events are responsible for an observed outcome. Answering them requires identifying the relevant objects, recovering the temporal order of their interactions, and relating a target event to preceding motion or collisions. The candidate answers refer to possible explanatory events, so the model must distinguish the interactions that account for the outcome from other events in the video.

\paragraph{Predictive questions.} These questions ask what will happen after the observed video ends. The model must infer how the objects will continue moving and whether their subsequent trajectories will lead to the events described by the answer options. This category connects the visible history of motion and collisions with future physical interactions, testing whether the observed evidence supports an accurate continuation of the scene.

\paragraph{Counterfactual questions.} These questions ask how events would change under a specified intervention, such as removing an object from the scene. The model must identify the intervention target and reason about how the remaining objects would interact under the modified conditions. Because removing an object can alter subsequent collisions and motion, answering requires considering the consequences of the edit throughout the relevant event sequence.

\paragraph{Metrics.} We report per-option and per-question accuracy for each category and overall. Per-option accuracy is the fraction of answer options whose correctness is judged correctly. Per-question accuracy counts a question as correct only when every associated option is judged correctly. We use overall per-question accuracy as the primary metric. Overall scores are computed across all evaluated questions or options, respectively.

\subsection{ContPhy}
\label{app:contphy}

ContPhy~\citep{zheng2024contphy} is a benchmark for learning and reasoning about physical concepts from videos of continuum systems. It connects observed dynamics with questions about physical properties and the consequences of changing a scene. Its diverse materials and interactions support evaluation of reasoning across different physical systems. We evaluate \method{} on 600 videos containing 1,950 questions.

\paragraph{Physical-property questions.} These questions ask about physical attributes that govern the behavior of the objects or materials in a scene. The model must connect observable motion, deformation, or interaction responses with the underlying property being queried. Different scene categories provide different forms of evidence, making this category a test of whether visual dynamics can support judgments about the physical characteristics of the participating entities.

\paragraph{Predictive questions.} These questions ask how a physical system will evolve after the observed interval. The model must interpret the current configuration and motion, then anticipate the subsequent behavior of the materials and objects involved. Depending on the scene, relevant evidence may include deformation, contact, constrained movement, or flow. The requested answer concerns the future outcome of the system under its existing conditions.

\paragraph{Counterfactual questions.} These questions ask about the outcome of changing a condition in the observed scene. The model must interpret the specified change, identify the affected entities, and reason about how their interactions would unfold in the modified setting. Across the material families, this requires connecting an intervention with changes in motion, deformation, or flow and evaluating the resulting outcome described by the question.

\paragraph{Goal-driven questions.} These questions specify a desired physical outcome and ask which intervention would produce it. The model must relate each candidate change to its consequences for the system and select the option that satisfies the stated goal. Reasoning therefore proceeds from a target outcome to a suitable change in the scene, using the observed configuration and dynamics to assess the available choices.

\paragraph{Cloth scenes.} Cloth scenes feature flexible surfaces that bend and deform as they move and interact with other objects. Their evolution depends on the initial configuration, material response, and contact geometry. Understanding these scenes requires following changes in surface shape and spatial relationships over time. They provide visual evidence for questions about material behavior, subsequent deformation, and the effects of changing the physical setup.

\paragraph{Rope scenes.} Rope scenes feature connected flexible structures whose motion is constrained by their geometry and attachments, including interactions with pulleys and attached objects. Movement in one part of the system can affect other connected parts. Reasoning about these scenes requires tracking these relationships and understanding how forces and constraints shape the resulting motion, supporting questions about physical properties and the consequences of interventions.

\paragraph{Soft-ball scenes.} Soft-ball scenes feature deformable bodies whose shapes change during motion and contact. Their responses to interactions provide evidence about material behavior and influence subsequent trajectories. Understanding these scenes requires relating object-level movement to visible deformation, including how the body responds around contact. The questions use this evidence to probe physical attributes, future outcomes, and changes induced by modifying the scene conditions.

\paragraph{Fluid scenes.} Fluid scenes feature flowing materials that interact with containers, boundaries, and obstacles. Their behavior is expressed through changes in spatial distribution and the movement of material between regions. Reasoning requires following the evolving flow and relating it to scene geometry and physical conditions. These scenes support questions about material properties, future distributions, and how changes to the environment affect the resulting flow.

\paragraph{Metrics.} We report question-answering accuracy for each question category and overall. Category accuracy is the number of correctly answered questions divided by the number of questions in that category. Overall accuracy is the total number of correctly answered questions divided by 1,950, so each evaluated question contributes equally to the final score.

\subsection{Real-World}
\label{app:real-world}

We collect a real-world dataset using consumer-grade cameras to evaluate physical reasoning from recorded interactions. The dataset contains 60 videos and 60 questions across three scene categories, with 20 videos and 20 questions per category. It focuses on two challenging reasoning tasks: counterfactual reasoning and prediction. Billiards scenes provide counterfactual questions, while bouncing-ball and toy-car scenes provide predictive questions.

\paragraph{Billiards scenes.} These scenes capture balls moving and colliding on a billiards table. Counterfactual questions ask how the interactions would change under a specified intervention, such as removing a ball. Answering requires identifying the relevant balls, interpreting their observed motion, and reasoning about how the intervention changes subsequent collisions. This category evaluates the connection between object-level interventions and the resulting sequence of physical events.

\paragraph{Bouncing-ball scenes.} These scenes capture the motion of bouncing balls and their interactions with surrounding surfaces. Predictive questions ask about physical outcomes after the observed motion. The model must connect the ball's trajectory and contact response with its subsequent movement, accounting for changes in direction caused by interactions. This category evaluates whether observed dynamics support predictions of future motion and contact events.

\paragraph{Toy-car scenes.} These scenes capture toy cars moving through a physical environment and interacting with other scene elements. Predictive questions concern the subsequent motion or interactions of the cars. Answering requires identifying the relevant vehicles, interpreting their trajectories, and relating their movement to the surrounding geometry. This category evaluates prediction from real object motion under the spatial constraints of the recorded scene.

\paragraph{Ground-truth acquisition.} Each scene is recorded in full from multiple synchronized viewpoints. For predictive questions, the reference answers are determined from the future portion of the complete recording, with complementary views used to recover the subsequent trajectory and resolve occlusions. Counterfactual outcomes are not directly observed; we therefore retain only questions for which human observers can determine the answer with high confidence from the multi-view recordings. In particular, cases involving unresolved spatial ambiguity or viewpoint occlusion are excluded.

\paragraph{Metrics.} We report per-question accuracy for each scene category and overall. Each category score is the number of correctly answered questions divided by 20, and the overall score is the total number of correct answers divided by 60. Since the categories contain equal numbers of questions, overall accuracy also equals the mean of the three category accuracies.

%% file: sections/appendix_results.tex
\section{More Experimental Results}
\label{app:more-results}

\subsection{Detailed Real-World Results}
\label{app:real-world-results}

Table~\ref{tab:real-world-detailed} reports detailed accuracy on the 60 real-world videos and questions. In addition to per-question accuracy, we report per-option accuracy for billiards and toy-car tasks. Each bouncing-ball question requires one yes/no judgment. Overall per-judgment accuracy aggregates the 80 billiards options, 20 bouncing-ball judgments, and 60 toy-car options, for 160 judgments in total.

\begin{table}[htbp]
\caption{Detailed real-world accuracy (\%). Each scenario contains 20 videos and 20 questions. Q: per-question; O: per-option; J: per-judgment. Counts are shown in parentheses. Per-question accuracy requires all associated option judgments to be correct. CF: counterfactual; Pred.: predictive. Bold denotes the best result in each column.}
\label{tab:real-world-detailed}
\centering
\footnotesize
\setlength{\tabcolsep}{2.7pt}
\renewcommand{\arraystretch}{1.08}
\begin{tabular*}{\textwidth}{@{\hspace{6pt}\extracolsep{\fill}}l*{7}{c}@{\hspace{6pt}}}
\toprule
\raisebox{-7pt}[0pt][0pt]{Model} & \multicolumn{2}{c}{Billiards (CF)} & Bouncing ball & \multicolumn{2}{c}{Toy car (Pred.)} & \multicolumn{2}{c}{Overall} \\
\cmidrule(lr){2-3}\cmidrule(lr){4-4}\cmidrule(lr){5-6}\cmidrule(lr){7-8}
& Q (20) & O (80) & Pred. / Q (20) & Q (20) & O (60) & Q (60) & J (160) \\
\midrule
\rowcolor{ATWTableBand}[6pt][6pt]
\multicolumn{8}{@{\hspace{6pt}}l@{\hspace{6pt}}}{\textit{Foundation VLMs}} \\
GPT-5.5 & 35.00 & 61.25 & 45.00 & 50.00 & 65.00 & 43.33 & 60.63 \\
Gemini-3-Flash & 20.00 & 60.00 & 60.00 & 20.00 & 26.67 & 33.33 & 47.50 \\
\midrule
\rowcolor{ATWTableBand}[6pt][6pt]
\multicolumn{8}{@{\hspace{6pt}}l@{\hspace{6pt}}}{\textit{Training-Free Methods}} \\
\textbf{\method{} (Ours)} & \textbf{85.00} & \textbf{91.25} & \textbf{70.00} & \textbf{60.00} & \textbf{71.67} & \textbf{71.67} & \textbf{81.25} \\
\bottomrule
\end{tabular*}
\end{table}

\subsection{Additional Ablation Results}
\label{app:additional-ablations}

\begin{figure}[htbp]
\centering
\includegraphics[width=\textwidth]{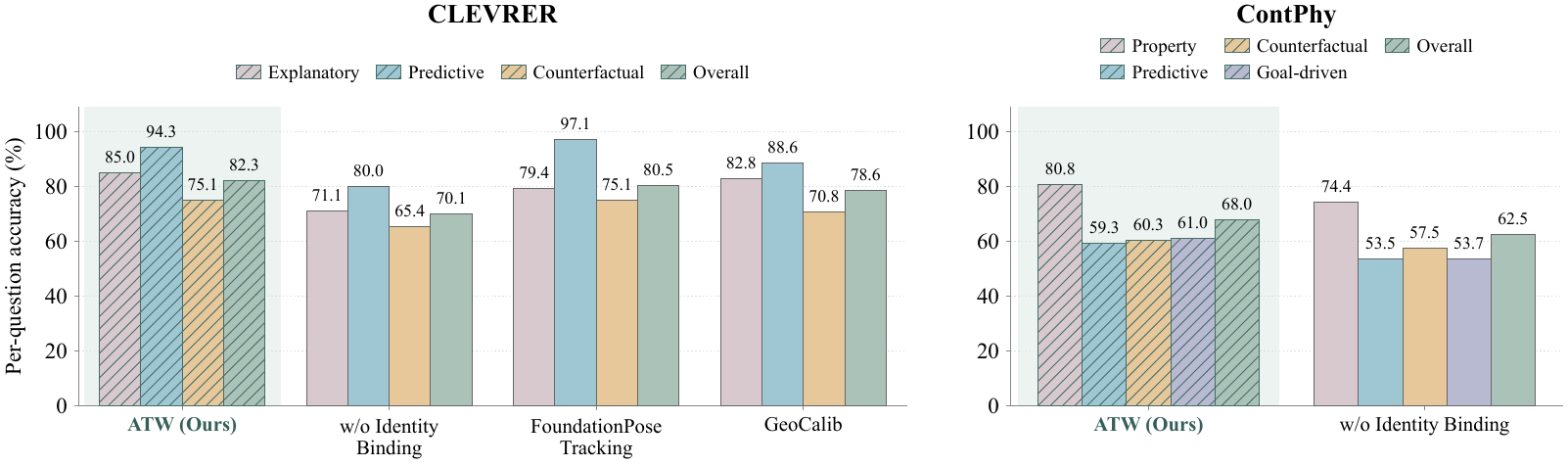}
\caption{Additional ablations on CLEVRER (left) and ContPhy (right). Bars show per-question accuracy for each reasoning category and overall, evaluated on 100 scenes per benchmark (435 CLEVRER questions and 325 ContPhy questions). Hatched bars denote full \method{}.}
\label{fig:additional-ablations}
\end{figure}

Figure~\ref{fig:additional-ablations} reports additional ablations on the same 100-scene subsets used in Section~4.3, with full \method{} included for comparison. The identity-binding ablation removes the appearance cards and identity-binding table.

\paragraph{Effects of identity binding and perception modules.}
Figure~\ref{fig:additional-ablations} shows that removing appearance cards and identity bindings reduces overall accuracy from 82.30\% to 70.11\% on CLEVRER and from 68.00\% to 62.46\% on ContPhy, with declines across every question category. These results highlight the importance of maintaining object identities when connecting visual observations, question references, and executable worlds. On CLEVRER, FoundationPose~\citep{wen2024foundationpose} tracking achieves 80.46\% overall accuracy and the highest predictive accuracy of 97.14\%, while matching \method{} on counterfactual questions. Its tracking benefits from an input object mesh and a rigid-body prior, making it effective for these rigid-object scenes. However, its rigid-pose formulation does not capture the evolving shapes of cloth, fluids, or soft bodies. We therefore use Track4World~\citep{lu2026track4world} for trajectory tracking across the diverse physical systems considered in \method{}. The GeoCalib~\citep{veicht2024geocalib} variant, which uses a learning-based single-image camera calibration method to estimate camera intrinsics and gravity direction, achieves 78.62\% overall accuracy. Full \method{} obtains the highest overall score, while different module choices produce distinct accuracy profiles across reasoning categories.

\subsection{Detailed Physical-Grounding Ablations}
\label{app:physical-grounding-details}

Tables~\ref{tab:contphy-system-identification} and~\ref{tab:clevrer-physical-grounding} provide the category-level results underlying the system-identification and physics-backend ablations. Overall accuracies are aggregated over all questions or options rather than averaged across categories.
These experiments are run independently of the agentic world-reasoning ablation.

\begin{table}[htbp]
\caption{Detailed system-identification ablation on the 100-scene ContPhy subset. Category columns report question-answering accuracy (\%). Random Search and CEM use equal simulation budgets.}
\label{tab:contphy-system-identification}
\centering
\small
\setlength{\tabcolsep}{4pt}
\renewcommand{\arraystretch}{1.10}
\begin{tabular*}{\textwidth}{@{\hspace{6pt}\extracolsep{\fill}}l*{5}{c}@{\hspace{6pt}}}
\toprule
Method & Property & Predictive & Counterfactual & Goal-driven & Overall \\
\midrule
\rowcolor{ATWTableBand}[6pt][6pt]
\multicolumn{6}{@{\hspace{6pt}}l@{\hspace{6pt}}}{\textit{System Identification (Warp Fixed)}} \\
Default parameters & 70.40 & 39.53 & 56.16 & 80.49 & 60.31 \\
Random Search & 70.40 & 51.16 & 63.01 & 46.34 & 60.62 \\
CEM (Ours) & 80.00 & 54.65 & 63.01 & 65.85 & 67.69 \\
\bottomrule
\end{tabular*}
\end{table}
\FloatBarrier

\paragraph{ContPhy trends.}
CEM achieves the strongest aggregate performance among the three configurations. Relative to Random Search, it improves physical-property, predictive, goal-driven, and overall accuracy, while matching counterfactual accuracy at 63.01\%. These results indicate that CEM-based system identification improves aggregate reasoning performance, with category-specific differences.

\begin{table}[htbp]
\caption{Detailed system-identification and physics-backend ablations on the 100-scene CLEVRER subset. Accuracy is reported in percent (Q: per-question; O: per-option), and the final column reports median trajectory RMSE in meters. The CEM and Warp + CEM rows are the same shared configuration.}
\label{tab:clevrer-physical-grounding}
\centering
\footnotesize
\setlength{\tabcolsep}{1.2pt}
\renewcommand{\arraystretch}{1.12}
\begin{tabular*}{\textwidth}{@{\hspace{3pt}\extracolsep{\fill}}l*{9}{c}@{\hspace{3pt}}}
\toprule
\raisebox{-7pt}[0pt][0pt]{Method}
& \multicolumn{2}{c}{Explanatory}
& \multicolumn{2}{c}{Predictive}
& \multicolumn{2}{c}{Counterfactual}
& \multicolumn{2}{c}{Overall}
& \raisebox{-7pt}[0pt][0pt]{\shortstack{Median trajectory\\RMSE (m)}} \\
\cmidrule(lr){2-3}\cmidrule(lr){4-5}\cmidrule(lr){6-7}\cmidrule(lr){8-9}
& Q & O & Q & O & Q & O & Q & O & \\
\midrule
\rowcolor{ATWTableBand}[6pt][6pt]
\multicolumn{10}{@{\hspace{3pt}}l@{\hspace{3pt}}}{\textit{System Identification (Warp Fixed)}} \\
Default parameters & 25.56 & 50.62 & 97.14 & 98.57 & 1.08 & 50.23 & 26.67 & 55.11 & 0.292053 \\
Random Search & 41.67 & 64.77 & 91.43 & 92.86 & 27.03 & 69.18 & 43.45 & 69.49 & 0.034992 \\
CEM (Ours) & 86.11 & 94.31 & 94.29 & 97.14 & 75.14 & 91.22 & 82.76 & 93.19 & 0.007655 \\
\midrule
\rowcolor{ATWTableBand}[6pt][6pt]
\multicolumn{10}{@{\hspace{3pt}}l@{\hspace{3pt}}}{\textit{Physics Backends (CEM Fixed)}} \\
PhysMind model + CEM & 58.33 & 76.15 & 94.29 & 95.71 & 33.51 & 72.88 & 53.56 & 76.58 & 0.059138 \\
MuJoCo + CEM & 68.89 & 83.38 & 91.43 & 94.29 & 71.89 & 87.52 & 73.79 & 86.31 & 0.016017 \\
Warp + CEM (Ours) & 86.11 & 94.31 & 94.29 & 97.14 & 75.14 & 91.22 & 82.76 & 93.19 & 0.007655 \\
\bottomrule
\end{tabular*}
\end{table}
\FloatBarrier

\paragraph{CLEVRER trends.}
Across the system-identification variants, trajectory RMSE decreases from default parameters to Random Search and then to CEM, while overall per-question and per-option accuracy increase in the same order. The largest reasoning gains appear in explanatory and counterfactual questions, while predictive accuracy remains high across configurations. With CEM fixed, the comparison from the PhysMind model through MuJoCo to Warp shows the same broad relationship between lower trajectory error and higher overall accuracy. Together, these trends show that both system identification and the physical backend determine the quality of the executable world used for downstream reasoning.

%% file: sections/appendix_prompts.tex
\section{System Prompts}
\label{app:system-prompts}

This section presents the principal system instructions used in \method{}. We organize them according to the main stages of the framework: world-representation routing, object discovery and segmentation, world probing, and evidence-based answer generation.

\begingroup
\definecolor{PromptBoxBackground}{gray}{0.95}
\definecolor{PromptBoxBorder}{gray}{0.72}
\newcommand{\promptsummary}[2]{%
  \par\medskip\noindent
  \setlength{\fboxsep}{7pt}%
  \fcolorbox{PromptBoxBorder}{PromptBoxBackground}{%
    \parbox{\dimexpr\linewidth-2\fboxsep-2\fboxrule\relax}{%
      \small\textbf{#1}\par\smallskip
      #2%
    }%
  }%
  \par\medskip
}

\promptsummary{World-representation routing.}{%
Inspect the complete video together with all associated questions, but do not answer them at this stage. First identify whether the main interactions involve free collision and support, constrained relative displacement, or another form of material motion. Examine depth ordering and outline truncation during contact. Use a three-dimensional representation when occlusion makes a visible mask boundary different from the physical contact surface, or when folding, covering, and other out-of-plane states cannot be described by planar outlines. Use a planar representation when the relevant connections, endpoints, displacements, and boundaries provide sufficient interaction geometry. Base the decision on visible evidence rather than object names, dataset conventions, or rendering style, and report the observations and remaining uncertainty that support the route.}

\promptsummary{Object discovery and SAM 3 segmentation.}{%
Inspect the entire video and indexed frame samples to enumerate every distinct physical entity in the interaction workspace. Include stationary obstacles, supports, and dividers that constrain the experiment, while excluding continuous environment surfaces, shadows, reflections, annotations, and rendering artifacts. Keep same-category instances separate and assign each entity a stable opaque identity. Classify its physical type as rigid, soft body, cloth, or rope from temporal evidence, distinguishing real deformation from occlusion, viewpoint change, and rigid rotation. For each entity, generate short image-grounded descriptions and select a seed frame that prioritizes a complete in-frame outline, visibility, and identity clarity. Accept a SAM 3 candidate only when it covers the intended foreground entity and excludes neighboring objects and background. If no candidate is adequate, refine the description or reselect the seed frame while preserving the target identity.}

\promptsummary{World probing (shared instructions).}{%
Investigate the literal question before producing an answer and treat its options independently. Match the referenced entities, world, intervention, and time interval. Missing evidence is unknown rather than false, so inspect returned results before deciding that the evidence is sufficient. Use only the tools exposed by the selected route and issue one to six independent tool calls in a round. A rollout identifier must come from an earlier successful operation, and subsequent queries must read from the corresponding world. Follow returned guidance when evidence is missing and do not repeat an unchanged failed request. Termination or video-fallback requests are issued alone. Because the answer stage receives compact textual evidence, preserve every necessary visual observation together with the evidence reference from which it was obtained.}

\promptsummary{Evidence-based answer generation.}{%
Answer the literal question using only its retained evidence. Match the referenced entities, intervention, and time window, and handle negated statements explicitly. Evaluate each relevant claim independently: unknown evidence does not imply a negative decision, and a claim that an event did not occur requires adequate coverage of the requested interval. Cite the evidence supporting the answer and provide a short factual reason without repeating the full tool history. Return one object that follows the specified output contract. If the retained evidence cannot determine the answer, return an explicit unresolved result that identifies the missing evidence rather than inventing a conclusion.}
\endgroup